\documentclass[letterpaper, 10 pt, conference]{ieeeconf}  

\IEEEoverridecommandlockouts                              

\usepackage{hyperref}
\usepackage{graphicx}
\usepackage{amssymb}
\usepackage{amsmath}
\usepackage{multirow}
\usepackage{threeparttable}
\usepackage{float}
\usepackage{tabularx}
\usepackage{booktabs}

\title{\LARGE \bf
DyGEnc: Encoding a Sequence of Textual Scene Graphs\\ to Reason and Answer Questions in Dynamic Scenes
}

\author{Sergey Linok$^{1,\dag}$, Vadim Semenov$^{1}$, Anastasia Trunova$^{1}$, Oleg Bulichev$^{1,2}$, Dmitry Yudin$^{1,3}$ 
\thanks{$^{1}$Center for Cognitive Modeling, Moscow Institute of Physics and Technology, Dolgoprudny, Russia}
\thanks{$^{2}$Innopolis University, Tatarstan, Russia}
\thanks{$^{3}$AIRI, Moscow, Russia}
\thanks{$^{\dag}$Corresponding author: linok.sa@phystech.edu}
}

\begin{document}

\maketitle
\thispagestyle{empty}
\pagestyle{empty}

\begin{abstract}
The analysis of events in dynamic environments poses a fundamental challenge in the development of intelligent agents and robots capable of interacting with humans. 
Current approaches predominantly utilize visual models. However, these methods often capture information implicitly from images, lacking interpretable spatial-temporal object representations. 
To address this issue we introduce DyGEnc - a novel method for Encoding a Dynamic Graph. 
This method integrates compressed spatial-temporal structural observation representation with the cognitive capabilities of large language models. 
The purpose of this integration is to enable advanced question answering based on a sequence of textual scene graphs. 
Extended evaluations on the STAR and AGQA datasets indicate that DyGEnc outperforms existing visual methods by a large margin of 15–25\% in addressing queries regarding the history of human-to-object interactions. 
Furthermore, the proposed method can be seamlessly extended to process raw input images utilizing foundational models for extracting explicit textual scene graphs, as substantiated by the results of a robotic experiment conducted with a wheeled manipulator platform. 
We hope that these findings will contribute to the implementation of robust and compressed graph-based robotic memory for long-horizon reasoning. 
Code is available at \url{github.com/linukc/DyGEnc}\footnote[4]{This work has been submitted to the IEEE for possible publication. Copyright may be transferred without notice, after which this version may no longer be accessible.}.

\end{abstract}

\section{INTRODUCTION}

Interpretable object maps for representing the surrounding environment for robots are an actively researched topic. These maps include descriptions — either explicit textual annotations or implicit representations in the form of extracted features — of scene elements along with their 3D positions and orientations, typically for subsequent utilization with large language models that facilitate logical analysis and reasoning of user's textual queries.

ConceptGraphs~\cite{gu2024conceptgraphs}, BBQ~\cite{linok2024beyond}, Search3D~\cite{takmaz2025search3d} and analogous approaches construct advanced graph structures from a sequence of positioned frames using fundamental visual models. This enables the identification of objects of interest through arbitrary text queries that specify diverse inter-object spatial relationships. HOV-SG~\cite{werby2024hierarchical} and Clio~\cite{maggio2024clio} employ a multi-level hierarchy to represent large interior spaces as layered graphs (e.g. floors, rooms), with each node preserving its unique features. This approach substantially narrows the search context for text queries and facilitates scaling the knowledge maps of intelligent agents to extensive areas.

\begin{figure}[t]
  \centering
  \includegraphics[width=1.0\columnwidth]{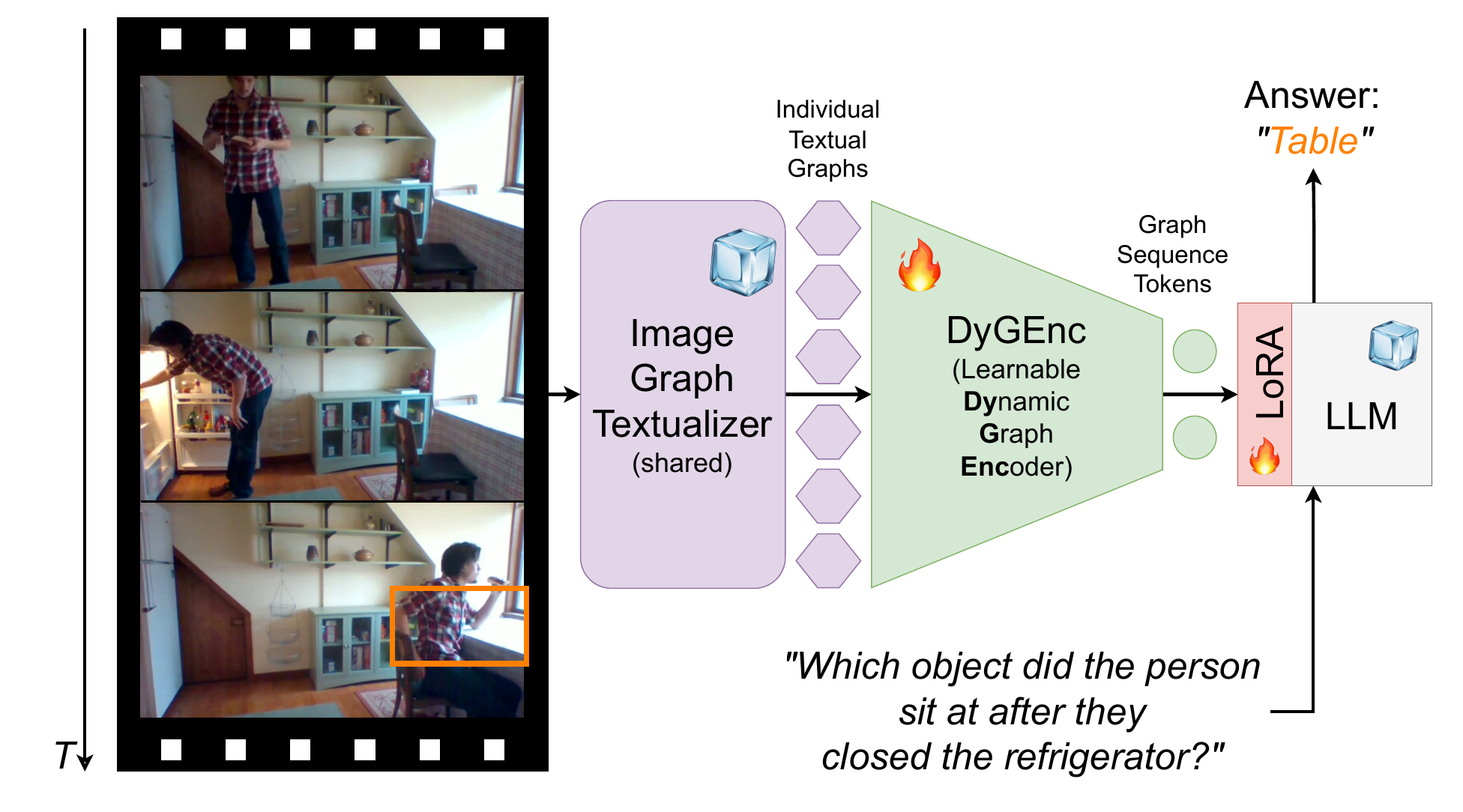}
  \caption{DyGEnc compactly encodes a dynamic graph (sequence of textual scene graphs) of a changing environment in a few tokens. The resulting representation is then utilized by an aligned large language model for situated logical reasoning and question answering.}
  \label{fig:1_graphical_abstract}
\end{figure}

The presented methods operate under the assumption that the observed environment is static, which significantly limits their potential deployment in real-world settings where variability is a key attribute. PSG4D~\cite{yang20234d} incorporates an object tracking procedure, allowing an observation to be represented as a graph of objects with edges that depend on the query time. However, if the resulting representation is entirely conveyed in textual form to the context of a large language model, it can lead to hallucinations during logical reasoning due to the large volume of information generated by continuous changes. G-Retriever~\cite{he2025g} advances the idea of encoding single graph representation~\cite{perozzi2024let,selvamcan}, enabling a concise and implicit depiction of the scene description in the form of specialized input tokens with high compression rate and without information quality drop. To address the context limit for dynamic scenes, we propose the DyGEnc method, which extends the encoding concept to sequences of graphs (dynamic graph), as illustrated in Fig.~\ref{fig:1_graphical_abstract}.

Thus, our key contributions are as follows: 
\begin{enumerate} 
    \item DyGEnc architecture for encoding sequence of textual scene graphs based on a parameter-efficient fine-tuning of a large language model (Sec.~\ref{sec:method});
    \item A comprehensive analysis of DyGEnc components and their impact on the model performance, evaluated on STAR~\cite{wu2024star} and AGQA~\cite{grundemclaughlin2022agqa20updatedbenchmark} benchmarks (Sec.~\ref{sec:ablation});
    \item A practical approach for deploying DyGEnc for real-world robotic applications by leveraging foundational models for extracting textual scene graphs from a sequence of images (Sec.~\ref{video_inference}). Code is available at \url{github.com/linukc/DyGEnc}.
\end{enumerate}

\section{RELATED WORK}
\subsection{Dynamic Scene Graph Generation}

We define Dynamic Scene Graph (DSG) as a sequence of graphs in which the connections are characterized not only by spatial relations between objects but also by action connections between moving actors and objects. The prediction of such graphs from sensory data (primarily image sequences) has been extensively researched. Two broad categories of modern methods can be identified: firstly, end-to-end trainable approaches, and secondly, graph construction based on foundational models.

The advent of trainable methods was precipitated by the emergence of manually labeled graph datasets such as Visual Genome~\cite{krishna2017visual}, GQA~\cite{hudson2019gqa}, PSG~\cite{yang2022panoptic}, Action Genome~\cite{ji2020action} and STAR~\cite{wu2024star}. Today, a vast set of diverse approaches exists, with some of the most state-of-the-art being PSG~\cite{yang2022panoptic}, PVSG~\cite{yang2023panoptic}, EGTR~\cite{im2024egtr}, Reltr~\cite{cong2023reltr}, and OED~\cite{wang2024oed} transformer image-based graph predictors.

In the second group of methods, the utilization of foundational models~\cite{ravi2024sam,liu2024llavanext,cheng2024yolo}, as well as modern large language models with visual input (vLLM)~\cite{achiam2023gpt,young2024yi,liu2024nvila,bai2025qwen2}, is worthy of particular attention. These models are actively employed to generate uprising synthetic graph annotations~\cite{GBC2024,zhang2024provision,kim2024llm4sgg}. The second group of approaches exhibits better generalization on new data than trainable methods on existing graph datasets, which are limited in their diversity — a key factor for the proposed algorithm's performance across a wide range of possible scenarios — but demand human-in-a-loop to correct hallucinations and verify output.

\subsection{Video Question Answering (VQA) with DSG}

In recent years, Video Question Answering (VideoQA) has emerged as one of the most rapidly developing research areas at the intersection of computer vision, natural language processing, and multimodal learning. However, as recently substantiated by studies in the field~\cite{yang2024thinking,zou2024seconds,li2025visual}, existing vLLMs encounter difficulties in accumulating perceived data in a meaningful inner representation due to their implicit encoding of input images. This feature plays a particularly critical role in dynamic scene understanding, where the relationships between entities are in constant flux. That is why alternative architectures are being developed, incorporating analogs of graph-based hierarchical representations.

The authors~\cite{cherian20222} propose to represent video as a (2.5+1)D scene graph, where each node has spatio-temporal coordinates. To construct the graph, they employ detection and tracking models pre-trained on a specific domain and train their transformer model for question answering. The creators~\cite{rodin2024action} assembled their graph dataset and trained a model to generate Egocentric Action Scene Graphs for video representation, while logical reasoning is performed by passing the information into the context of a large language model. HyperGLM~\cite{nguyen2024hyperglm} proposes a Video Scene HyperGraph, where hyperedges are depicted as polygons, encapsulating interactions through chains of relationships. STEP~\cite{qiu2024step} presents a procedure for fine-tuning an existing video model by applying symbolic structure induction in the SpatioTemporal Scene Graph and stepwise graph-driven rationale learning.

In DyGEnc we propose to encode a sequence of textual scene graphs utilizing a graph neural network to preserve not only semantics but also the relationships between observed objects at each unique moment and sequence encoder for hidden representations compression. Moreover, DyGEnc employs parameter-efficient fine-tuning of a large language model, leveraging its inherent potential for logical reasoning in the text modality over implanted DSG tokens.

\section{METHOD}
\label{sec:method}

\begin{figure*}[t]
  \centering
  \includegraphics[width=2.0\columnwidth]{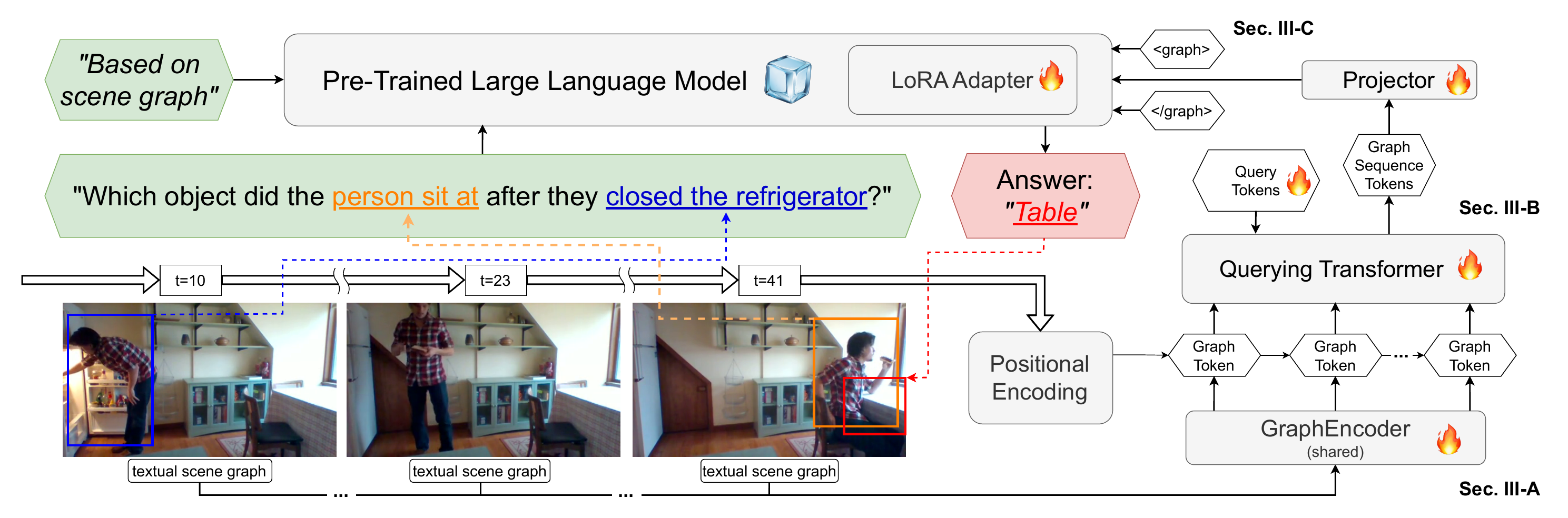}
  \caption{Overview of the DyGEnc pipeline. Given a dynamic scene graph - a sequence of textual scene graphs, where nodes and edges carry attributes encoded by a pre-trained text encoder, we first pass each encoded graph through a graph neural network to generate an aggregated graph token. To preserve temporal information, each graph token is enriched with a positional encoding. Then Q-Former module is applied to capture temporal relations, producing a compact sequence representation in query tokens. Finally, a multilayer perceptron projects these tokens into a large language model’s embedding space, with special tokens marking the start and end of the graph soft prompt. Thus LLM can ground its reasoning based on sensory input.}
  \label{fig:2_method}
\end{figure*}

A textual scene graph is a graph derived from an image where nodes and edges possess textual attributes and represent objects and their relations. Formally, it can be defined as $G=(V,E)$, where $V$ and $E$ represent the sets of nodes and edges, respectively. Dynamic graph is a sequence of G.

\subsection{Graph Encoding}
\label{sec:graph_encoding}

Consider $x_n$ as the text attributes of node $n$. Utilizing a pre-trained text encoder $LM$, we apply it to $x_n$, yielding the representation $z_n$:
\begin{equation}
    \label{eq:1}
    z_n = LM(x_n) \in \mathbb{R}^d,
\end{equation}
where $d$ denotes the dimension of the output vector. Similar preprocessing steps are applied to edges. We utilize a pretrained ModernBert~\cite{warner2412smarter} base version with 149M parameters as $LM$.

Additionally, for each node $n$, laplacian positional encoding $d_{lpe} \in \mathbb{R}^{4}$~\cite{dwivedi2023benchmarking} is added to the text encoder’s dimensionality $d$.

Then latent representation $G_z=(V_z,E_z)$ used to encode graph structure with a graph neural encoder $GNN$:
\begin{equation}
    \label{eq:2}
    h_g = F_{aggr}(GNN_{\theta_1}(G_z)) \in \mathbb{R}^{d_g}.
\end{equation}
Here, $F_{aggr}$ denotes the aggregation scheme, and $d_g$ is the output dimension of the graph encoder. For the $F_{aggr}$ we use mean pooling and for \textit{GNN} — GraphTransformer~\cite{shi2020masked} with 21M parameters.

\subsection{Sequence Encoding}

To preserve original timeline, we extend each graph token $h_G$ with a positional encoding vector $p_{tpe} \in \mathbb{R}^{d_g}$:
\begin{equation}
    \label{eq:3}
    \hat{h}_g = h_g + p_{tpe} \in \mathbb{R}^{d_g},
\end{equation}
where $t$ is an graph index in a sequence. We utilize Rotary Positional Encoding~\cite{su2024roformer} for temporal encoding. Ablation study of different positional encoding sceme can be found in Sec.~\ref{ablation_pos_encoding}.

To encode temporal relations we utilize sequence encoder \textit{SE} with Q-Former~\cite{li2023blip} architecture of cross-attention decoder transformer with 2 layers and 4 cross-attention heads each, resulting in 19M parameters $\theta_2$. For a given sequence of $m$ scene graph tokens $\hat{h}_g$ ($M \in \mathbb{R}^{m \times d_g}$), cross-attention to $k$ learnable query tokens $K \in \mathbb{R}^{k \times d_g}$ ($k\leq m$) are appliend to gather sequence information in compact latent representation of dynamic graph:
\begin{equation}
    \label{eq:3}
    \hat{h}_{dg} = SE_{\theta_2}(M, K) \in \mathbb{R}^{k \times d_g},
\end{equation}
Fixed number of output tokens $K$ and theoretically unconstraned number of $M$ in potential scene graph sequence allows to apply sequence encoding to arbitrary sequence len while preserving constant context size for a large language model. Ablation study on number of learnable query tokens $K$ with respect to parameters of the Q-Former can be found in Sec.~\ref{ablation_q_tokens}.

\subsection{LLM Finetuning}

To align compressed tokens, describing sequence of graphs, we project $K$ to vector space of the LLM by incorporate a multilayer perceptron $MLP$:
\begin{equation}
    \label{eq:4}
    h_{llm} = MLP_{\theta_3}(\hat{h}_{dg}) \in \mathbb{R}^{k \times d_{llm}},
\end{equation}
where $d_{llm}$ is the dimension of the LLM’s hidden embedding.

The final stage involves generating the answer $A$ given the list of latent dynamic graph tokens $h_{llm}$, acting as a soft prompt, and question $Q$. These concatenated inputs are fed through the self-attention layers of a pretrained frozen $LLM$ with parameters $\theta_4$:
\begin{equation}
    \label{eq:5}
    A = LLM_{\theta_4}(Q, h_{llm}).
\end{equation}
While most of $\theta_4$ are frozen, part of the weights $\hat{\theta_4}$ are updated with parameter-efficient training alongside with the graph tokens $h_{llm}$ receiving gradients, enabling the optimization of the parameters $\theta_3$ of the projection layer, Q-Former $\theta_2$ and graph encoder $\theta_1$ through backpropagation. More technical and implementation details can be found in Sec~\ref{finetune}.

\section{EXPERIMENTS}
\label{sec:experiments}

\subsection{Datasets}
\subsubsection{STAR}
Benchmark~\cite{wu2024star} for situated reasoning is built upon the 9K real-world videos of human actions and surrounding environments in daily-life scenes. Annotation consists of 60K situated questions divided into four types, including interaction, sequence, prediction, and feasibility, for 22K video clips labeled with scene graphs. Lengths of graph sequences have a positively skewed distribution with a median of 20, interquartile range of 15, 5th percentile equals 7, and 95th percentile equals 46.

\subsubsection{AGQA}
Action Genome Question Answering benchmark~\cite{grunde2021agqa} for compositional spatio-temporal reasoning consists of visual events that are a composition of temporal actions involving actors interacting with objects. We use the latest (second) version of the benchmark~\cite{grundemclaughlin2022agqa20updatedbenchmark} that contains 9.6K unique scene graph sequences with annotations from real-life videos with a positivly skewed distribution with median sequence lengths of 28, interquartile range of 20, 5th percentile equals to 10, 95th percentile equals to 60. 2.27M balanced question answer pairs are generated from more than 30 diverse templates covering reasoning, structure and semantic understanding in 16 subcategories.

\subsection{Evaluation metrics}

To evaluate answer quality, we use \textit{Accuracy} as a primary metric to be in alignment with previous research, borrowing metrics from the corresponding publications. Under this metric, our prediction is considered correct if the dataset ground true answer contains the response generated by the model. Also, for experiments with DSG construction on our dataset (Sec.~\ref{video_inference}), where the model works mostly with textual scene graphs from out-of-training distribution and generated answer can be semantically close, but differ from answer label, we extend the evaluation metric by \textit{BLEU}, \textit{METEOR}, and \textit{BERTScore} from the HuggingFace Evaluate package.

\begin{table}[t]
    \setlength{\tabcolsep}{3pt}
    \centering
    \scriptsize
    \caption{Ablation of temporal positional encoding \\ on STAR~\cite{wu2024star} QA validation split}
    \label{tab:ablation_pos_enc}
    \begin{tabularx}{0.45\textwidth}{c c | c c c c | c}
        \toprule
        Size & Encoding & Interaction & Sequence & Prediction & Feasibility & Average \\
        \midrule
        3B & \multirow{2}{*}{TE} & \underline{0.96} & \underline{0.89} & 0.7 & 0.63 & \underline{0.88} \\
        8B &                     & 0.92 & 0.82 & 0.68 & 0.55 & 0.82 \\
        \midrule
        3B & \multirow{2}{*}{APE} & \textbf{0.97} & 0.88 & \textbf{0.77} & \textbf{0.69} & \textbf{0.89} \\
        8B &                      & 0.95 & 0.85 & \underline{0.76} & 0.65 & 0.86 \\
        \midrule
        3B & \multirow{2}{*}{RoPE} & \textbf{0.97} & \textbf{0.9} & \textbf{0.77} & 0.65 & \textbf{0.89} \\
        8B &                      & \underline{0.96} & 0.87 & \textbf{0.77} & \underline{0.67} & \underline{0.88} \\
        \bottomrule
    \end{tabularx}
\end{table}

\begin{table}[t]
    \centering
    \scriptsize
    \setlength{\tabcolsep}{3pt}
    \caption{Ablation of Q-Former query token numbers \\ on STAR~\cite{wu2024star} QA validation split}
    \label{tab:ablation_q_former}
    \begin{tabularx}{0.45\textwidth}{c c | c c c c | c}
        \toprule
        Size & \begin{tabular}{c}Num\\Tokens\end{tabular} & Interaction & Sequence & Prediction & Feasibility & Average \\
        \midrule
        3B & \multirow{2}{*}{1} & 0.97 & 0.9 & 0.77 & 0.65 & 0.89 \\
        8B &                    & 0.96 & 0.87 & 0.77 & 0.67 & 0.88 \\
        \midrule
        3B & \multirow{2}{*}{2} & 0.97 & 0.9 & 0.78 & 0.7 & 0.9 \\
        8B &                    & 0.97 & 0.89 & 0.75 & 0.64 & 0.89 \\
        \midrule
        3B & \multirow{2}{*}{4} & 0.96 & 0.9 & 0.8 & 0.69 & 0.9 \\
        8B &                    & \textbf{0.99} & \underline{0.92} & 0.86 & 0.73 & 0.92 \\
        \midrule
        3B & \multirow{2}{*}{16} & \underline{0.98} & \textbf{0.94} & \textbf{0.91} & \textbf{0.79} & \textbf{0.94} \\
        8B &                     & \underline{0.98} & \underline{0.92} & \underline{0.89} & \underline{0.77} & \underline{0.93} \\
        \bottomrule
    \end{tabularx}
\end{table}

\begin{table}[t]
    \centering
    \scriptsize
    \caption{Ablation of DyGEnc components \\ on STAR~\cite{wu2024star} QA validaion split}
    \label{tab:components_on_star_val}
    \begin{tabularx}{0.45\textwidth}{l c | c c c c | c c}
        \toprule 
        Setup & Size & Int. & Seq. & Pred. & Feas. & Avg. & Compr.\\
        \midrule
        \textbf{Zero-shot} & & & & & & \\
        \multirow{2}{*}{-} & 3B & 0.35 & 0.26 & 0.38 & 0.34 & 0.3 & 1x \\
        & 8B & 0.34 & 0.28 & 0.4 & 0.35 & 0.32 & 1x \\
        \midrule
        \textbf{Fine-tuning} & & & & & & \\
        - & 3B & \textbf{1.0} & \textbf{1.0} & \textbf{0.98} & \textbf{0.95} & \textbf{0.99} & 1x \\
        \multirow{2}{*}{GE} & 3B & 0.96 & 0.92 & 0.89 & 0.78 & 0.92 & \underline{0.05x} \\
        & 8B & 0.96 & \underline{0.93} & \underline{0.92} & 0.82 & 0.93 & \underline{0.05x} \\
        \multirow{2}{*}{GE, TE} & 3B & 0.96 & 0.91 & 0.9 & 0.78 & 0.92 & \underline{0.05x} \\
        & 8B & 0.96 & \underline{0.93} & \underline{0.92} & \underline{0.84} & \underline{0.94} & \underline{0.05x} \\
        \multirow{2}{*}{GE, SE} & 3B & 0.97 & 0.91 & 0.69 & 0.59 & 0.89 & \textbf{0.03x} \\
        & 8B & \underline{0.98} & 0.88 & 0.68 & 0.58 & 0.87 & \textbf{0.03x} \\
        \multirow{2}{*}{GE, TE, SE} & 3B & 0.97 & 0.9 & 0.77 & 0.65 & 0.89 & \textbf{0.03x} \\
        & 8B & 0.96 & 0.87 & 0.77 & 0.67 & 0.88 & \textbf{0.03x} \\
        \bottomrule
    \end{tabularx}
\end{table}

\subsection{Implementation Details}
\subsubsection{Data Preprocessing}

Each sequence undergoes a preprocessing step that retains unique graphs. This can help significantly reduce the context for the model for observations in a low-dynamic environment. To preserve the temporal component — for example, to reason about time intervals and durations (Tab.~\ref{tab:ablation_agqa_test}) — the indices $t$ correspond to the indices of the graphs in the original sequence are preserved.

\subsubsection{LLM Finetuning}
\label{finetune}

For parameter-efficient LLM finetuning, a LoRA adapter~\cite{hu2022lora} is used with parameters \textit{r=8}, \textit{alpha=16}, and a dropout rate of 0.05, specifically targeting the $q\_proj$ and $v\_proj$ parts of the LLM's attention modules. In our work, we experiment with modern open Llama3~\cite{dubey2024llama} model family. We chose Llama 3.1-8B and Llama3.2-3B versions to meet the resource criteria of most potential application systems.

To adapt the language model for understanding the concept of graph tokens we add special tokens \textless graph\textgreater\ and \textless /graph\textgreater\ to represent the start and the end of dynamic graph latent representation, resulting with an input prompt: \textit{``Based on scene graph, \textless graph\textgreater\/$h_{llm}$\textless /graph\textgreater, Q"}.

We set AdamW optimizer with an initial learning rate at \textit{2e-5} and a weight decay of \textit{0.05}. The learning rate decays with a half-cycle cosine decay after the warm-up period of \textit{1} epoch. The batch size is \textit{32} and the number of epochs is set to \textit{5}. To prevent overfitting and ensure training efficiency, an early stopping mechanism is implemented with a patience setting to \textit{2} epochs. All experiments are done on a A100 80GB GPU. With such parameters, training on STAR takes approximetly one hour and near $\times10$ for AQGA. This is why we chose STAR for ablation study in Sec.~\ref{sec:ablation}. For both datasets we use same training hyperparameters.

\subsection{Ablation Study on STAR Benchmark}
\label{sec:ablation}

\begin{figure*}[t]
  \centering
  \includegraphics[width=2.0\columnwidth]{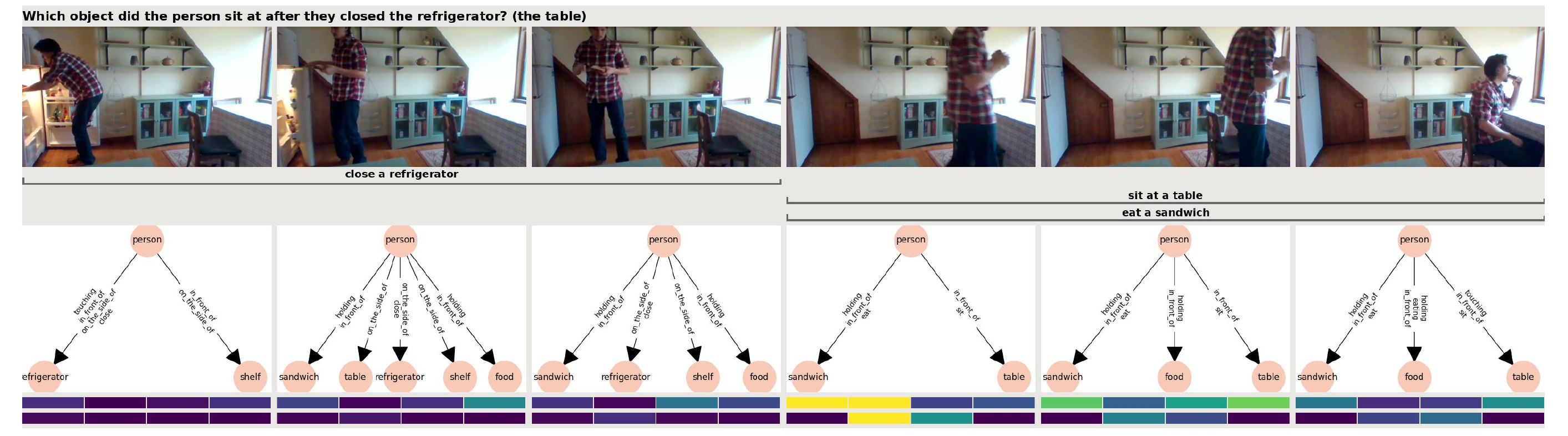}
  \caption{Example of cross-attention visualization from \textit{Q-Former} sequence encoder on STAR benchmark for the text query \textit{``Which object did the person sit at after they closed the refrigerator?"}. We draw cross-attention of 1 \textit{Q-Former} latent query token to each input graph embedding where 8 blocks represent 2 layer with 4 heads in each. Brighter color represents more model attention.}
  \label{fig:1_vis_attention}
\end{figure*}

\begin{figure*}[t]
  \centering
  \includegraphics[width=2.0\columnwidth]{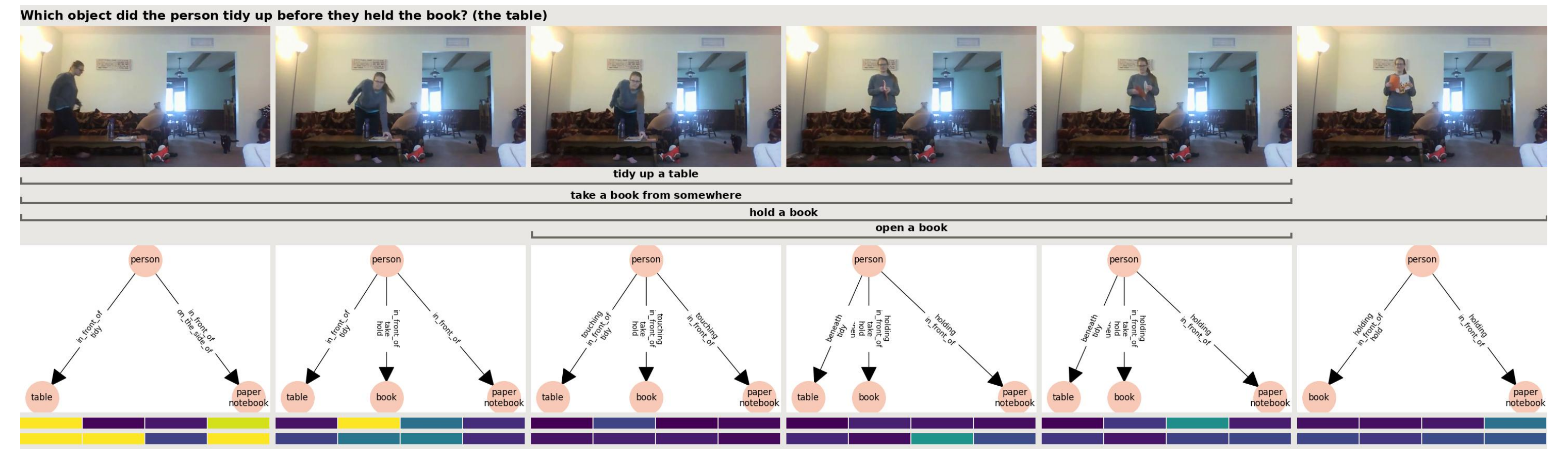}
  \caption{Example of cross-attention visualization from \textit{Q-Former} sequence encoder on STAR benchmark for the text query \textit{``Which object did the person throw before they held the dish?"}. We draw cross-attention of 1 \textit{Q-Former} latent query token to each input graph embedding where 8 blocks represent 2 layer with 4 heads in each. Brighter color represents more model attention.}
  \label{fig:2_vis_attention}
\end{figure*}

To better understand sequence encoding, we conduct ablation studies of key components: type of temporal positional encoding, which should preserve events time order, and \textit{Q-Former} hyperparameters search to understand to which degree we can compress dynamic scene graph tokens without significant loss of information for \textit{LLM}.

\subsubsection{Temporal Encoding}
\label{ablation_pos_encoding}

For temporal positional encoding we finetuned both LLM versions with three different approaches: Temporal Encoding~\cite{cong2023we} (\textit{TE}), Absolute Positional Encoding~\cite{vaswani2017attention} (\textit{APE}), Rotary Positional Encoding~\cite{su2024roformer} (\textit{RoPE}). For all runs, we set a number of \textit{Q-Former} query tokens equal to 1, because we are highly interested in the impact of encoding on extreme sequence compression level. Results in Table~\ref{tab:ablation_pos_enc} show that despite the close values of the \textit{Recall} metric, classical transformer positional encoders produce more consistent results for both model sizes than time-specific analog. For further experiments we use \textit{RoPe} approach as default.

\subsubsection{Number of Query Tokens}
\label{ablation_q_tokens}

We finetune both LLM models with different number of latent \textit{Q-Former} query tokens as a function of the fixed number of cross-attention layer (each layer with fixed number of heads queals to \textit{4}) under the hypothesis that as the degree of sequence compression increases, more attention parameters should be learned to effietefly handle data compression. Results in Table~\ref{tab:ablation_q_former} in general confirm our assumption. In the \textit{Prediction} and \textit{Feasibility} categories of the benchmark, we may see a significant drop in \textit{Recall} metric. However, these types of questions have highly biased answers that cannot always be logically deduced from dynamic graph. For the \textit{Interaction} and \textit{Sequence} categories where the model should understand events and their ordering, which is our main subject of study, the drop of metric with a compression rate increase is negligible. Thus, we set a number of latent \textit{Q-Former} query tokens equal to \textit{1}.

\subsubsection{DyGEnc Components Influence}

To validate the necessity of dynamic graph soft token representation, we first compare pre-trained Llama models between zero-shot and supervised runs, as shown in Tab.~\ref{tab:components_on_star_val}. For both setups, we textualize all graphs and concatenate them in one large corpus of text, which describes a sequence of graphs. Fine-tuning experiments show overwhelming superiority with extremely high convergence, illustrating that even a few samples of text descriptions are enough to give model context understanding. 

However, these achievement comes with a great cost of context size, even not allowing us to finetune the 8B model due to OOM. Here DyGEnc components help to drastically reduce tokens size as demonstrated in the last column of the table (\textit{Compr.} stands for \textit{Compression}). \textit{0.03x} degree of compression is achieved with the utilization of graph encoder (\textit{GE}) and sequence encoder (\textit{SE}), while temporal positional encoding (\textit{TE}) helps gain back some quality level.

\subsubsection{Evaluation Split}
\label{star_eval}

With the selected hyperparameters, we conduct a comparison with other methods using both variants of LLM. Results can be found in Tab.~\ref{tab:ablation_star_eval}. DyGEnc shows high-quality metrics, especially in the \textit{Interaction} and \textit{Sequence} categories compared to existing visual and visual-graph methods, utilizing observed sensory information only in a structural form of textual scene graphs sequence. 

In Fig.~\ref{fig:1_vis_attention} and Fig.~\ref{fig:2_vis_attention} we depict cross-attention to highlight that our method does not memorizes information during training, but learns how to attend to relevant frames and reason based on a sequence of scene graphs that represented as latent tokens.

\begin{table}[t]
    \centering
    \begin{threeparttable}
        \setlength\tabcolsep{1pt}
        \caption{Comparison with prior methods \\ on STAR~\cite{wu2024star} QA validation split}
        \label{tab:ablation_star_eval}
        \begin{tabularx}{0.45\textwidth}{l | c c c c | c}
            \toprule
            Method & Interaction & Sequence & Prediction & Feasibility & Average \\
            \midrule
            STEP~\cite{qiu2024step} & - & - & - & - & 0.4 \\
            Q-ViD~\cite{romero2024question} & 0.48 & 0.47 & 0.44 & 0.43 & 0.46 \\
            MIST~\cite{gao2023mist} & 0.56 & 0.54 & 0.54 & 0.45 & 0.51 \\
            SeViLA~\cite{yu2023self} & 0.64 & 0.71 & 0.63 & 0.62 & 0.65 \\
            ViLA~\cite{wang2024vila} & 0.7 & 0.7 & 0.66 & 0.62 & 0.67 \\
            VidF4~\cite{liang2024end} & 0.68 & 0.7 & 0.61 & 0.59 & 0.68 \\
            LRR~\cite{bhattacharyya2023look} & 0.74 & 0.71 & \underline{0.71} & \underline{0.65} & 0.71 \\
            \midrule
            \begin{tabular}{l}DyGEnc (ours)\\(Llama3.2-3B)\end{tabular} & \textbf{0.97} & \textbf{0.9} & \textbf{0.77} & 0.65 & \textbf{0.89} \\
            \begin{tabular}{l}DyGEnc (ours)\\(Llama3.1-8B)\end{tabular} & \underline{0.96} & \underline{0.87} & \textbf{0.77} & \textbf{0.67} & \underline{0.88} \\
            \bottomrule
        \end{tabularx}
  \end{threeparttable}
\end{table}

\begin{figure*}[t]
  \centering
  \includegraphics[width=2.0\columnwidth]{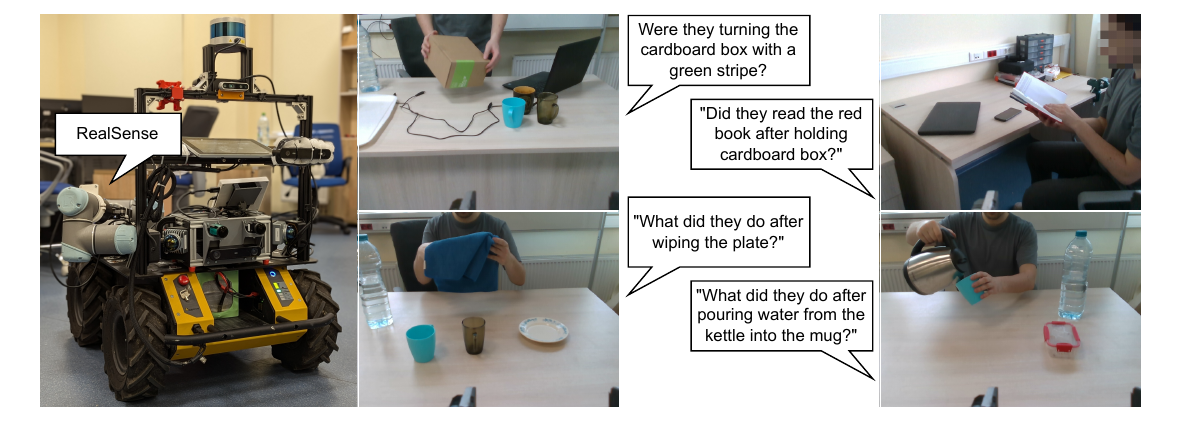}
  \caption{Illustration for a robotic experiment setup - to the left: mobile platform Husky with UR5 manipulator equipped to perform MOVE-AND-PICK task based on DyGEnc output, to the right - general scene overviews from our DRobot benchmark.}
  \label{fig:drobot}
\end{figure*}

\subsection{Evaluation on AQGA benchmark} 
\label{agqa_eval}

\begin{table}[t]
    \centering
    \begin{threeparttable}
        \scriptsize
        \setlength\tabcolsep{1.5pt}
        \caption{Comparison with prior methods \\ on AGQA2.0~\cite{grundemclaughlin2022agqa20updatedbenchmark} QA test split}
        \label{tab:ablation_agqa_test}
        \begin{tabularx}{0.45\textwidth}{l | c c c c c c c c | c c c}
            \toprule
            Method & \begin{tabular}{c}Obj.\\Rel.\end{tabular} & \begin{tabular}{c}Rel.\\Act.\end{tabular}& \begin{tabular}{c}Obj.\\Act.\end{tabular} & Sup. & Seq. & E. & Dur. & \begin{tabular}{c}Act.\\Rec.\end{tabular} & B. & O. & A.\\
            \midrule
            MIST~\cite{gao2023mist} & 0.52 & \underline{0.67} & \underline{0.69} & 0.42 & 0.67 & 0.6 & \textbf{0.55} & 0.2 & 0.58 & 0.51 & 0.54 \\
            GF~\cite{bai2023glance} & 0.55 & - & - & 0.45 & 0.53 & 0.59 & \underline{0.53} & 0.14 & 0.54 & 0.56 & 0.55 \\
            IPRM~\cite{jaiswal2025learning} & 0.58 & - & - & 0.48 & \textbf{0.76} & 0.62 & 0.51 & 0.2 & \underline{0.62} & 0.59 & 0.6 \\
            TGB~\cite{wang2024efficient} & 0.62 & 0.52 & 0.66 & 0.54 & 0.6 & 0.61 & 0.37 & 0.0 & - & - & 0.62\\
            DeST~\cite{lee2022learning} & 0.6 & \textbf{0.73} & \textbf{0.75} & 0.49 & \underline{0.74} & 0.63 & 0.6 & \underline{0.28} & \textbf{0.63} & 0.61 & 0.62\\
            \midrule
            \begin{tabular}{l}DyGEnc (ours)\\(Llama3.2-3B)\end{tabular} & \textbf{0.77} & 0.53 & 0.55 & \textbf{0.58} & 0.53 & \textbf{0.7} & 0.49 & \textbf{0.38} & \textbf{0.63} & \textbf{0.83} & \textbf{0.73} \\
            \begin{tabular}{l}DyGEnc (ours)\\(Llama3.1-8B)\end{tabular} & \underline{0.74} & 0.51 & 0.53 & \underline{0.55} & 0.52 & \underline{0.65} & 0.5 & \underline{0.28} & 0.6 & \underline{0.82} & \underline{0.71} \\
            \bottomrule
        \end{tabularx}
  \end{threeparttable}
\end{table}

AGQA is a most comprehensive benchmark with graph annotations with almost 2 million QA pairs. That's why we chose to benchmark our model only after hyperparameter ablations on the STAR dataset.
With the same settings as in Sec.~\ref{star_eval}, we trained both LLM models on the AQGA train split. For training, we limit the sequence length of unique graphs to 60 which preserves more than 95 percent of the original train data and allows us to discard samples that significantly affect the convergence process. This allows to stabilize the training process. It should be noted that we do not limit test split in the same manner.
Results of the evaluation can be found in Table~\ref{tab:ablation_agqa_test}. \textit{Obj.-Rel.}, \textit{Rel.-Act.}, \textit{Obj.-Act.}, \textit{Sup.}, \textit{Seq.}, \textit{Dur.}, \textit{E.} and \textit{Act.-Rec.} describe the question split and stand correspondingly for \textit{Object-Relationship}, \textit{Relationship-Action}, \textit{Object-Action}, \textit{Superlative}, \textit{Sequencing}, \textit{Duration}, \textit{Exists} and \textit{Activity Recognition}. \textit{B.} and \textit{O.} mean \textit{Binary} and \textit{Open} questions. \textit{A.} stands for \textit{All}. For all columns, we report \textit{Accuracy} metric. Our evaluation shows that compared to existing methods DyGEnc outperforms most existing works, especially on open-formulated (not binary) queries by a large margin, proving capabilities to distinguish unique text graph features and with reasoning capabilities to understand the semantics of question and context.

\subsection{DyGEnc Inference on Video}
\label{video_inference}

The experiments conducted on the previously described benchmarks utilized the available scene graph annotations. However, for the application of DyGEnc to real-world data, it is necessary to develop an algorithm capable of constructing textual scene graphs from a sequence of images. In Sec.~\ref{dsg_construction}, we introduce methodologies for generating textual scene graphs through the use of foundation models, and in Sec. ~\ref{subgraph_retrieval}, we present an algorithm designed for the extraction of subgraphs, aimed at reducing the input context to only those subgraphs that are relevant to the input text query.

\subsubsection{Subgraph Retrieval}
\label{subgraph_retrieval}

LLMs are known to hallucinate, meaning they generate incorrect or fabricated information. Without retrieval, an LLM might guess answers based on incomplete or incorrect context. Retrieval selects only the most relevant subgraph, significantly reducing the number of nodes, edges, and tokens processed. The retrieval step ensures that only relevant graph information is used, reducing the chance of incorrect responses. This speeds up inference time and makes it feasible for large-scale applications. In our practical robotic experiments we use G-Retrieval~\cite{he2024g} subgraph retrieval approach based on Prize-Collecting Steiner Tree algorithm~\cite{bienstock1993note_pcst} over user query embedded with the same text encoder as in Sec.~\ref{sec:graph_encoding} and graph embeddings.

\subsubsection{Ablation Study on DSG Construction}
\label{dsg_construction}

To generate textual scene graphs from images we compare three different approaches. The first approach utilizes Nvilda vLLM for image captioning and then uses Factual model specifically trained to extract triplets from the input text, describing an image. The second approach uses the same method to capture presents on the image, but uses GPT LLM to extract triplets from text. The last approach solely relies on GPT processing with a visual input to describe the image. For our experiments, we utilize the 4o-mini model. 

To compare these methods we constrct our DRobot benchmark. It consists of 10 scenes, each with the presence of some dynamic actions between humans and objects. Benchmark also persuites second important goal: show the use case of the DyGenc model on the sensory video input from the real robot. For this, we set up a robotic experiment in which a mobile-wheeled robot with a manipulator (depicted in Fig.~\ref{fig:drobot} to the left) has the task of moving to and picking an object of interest (a result of the answer). We created 50 questions following the template-based approach from the AQGA. It should be noted, that our environment has a major difference in object set between training distribution (examples are in Fig.~\ref{fig:drobot} to the right). For the experiments, we used pre-trained DyGenc-3B on the AGQA. Evaluation results in Tab.~\ref{tab:dsg_on_star_test} highlight, that with GPT's textual scene graphs, we can get higher metrics, but open-source analogs can also produce results without significant drops. We add a video attachment of the described experiment in the supplementary materials.

\begin{table}[t]
    \centering
    \setlength\tabcolsep{3pt}
    \caption{Ablation of SGD Constuction Algorithm \\ on our DRobot Dataset}
    \label{tab:dsg_on_star_test}
    \begin{tabularx}{0.45\textwidth}{l c | c c c c}
        \toprule
        Textualizer & \begin{tabular}{c}Graph \\ Constructor\end{tabular} & Acc. & BLEU & METEOR & BERTScore \\
        \midrule
        Nvila~\cite{liu2024nvila} & Factual~\cite{li2023factual} & 0.3 & 0.32 & 0.18 & \textbf{0.94} \\
        Nvila~\cite{liu2024nvila} & GPT~\cite{achiam2023gpt} & \textbf{0.34} & \underline{0.36} & \underline{0.2} & \textbf{0.94} \\
        \multicolumn{2}{c|}{GPT(v)~\cite{achiam2023gpt}} & \textbf{0.34} & \textbf{0.41} & \textbf{0.21} & \underline{0.92} \\
        \bottomrule
    \end{tabularx}
\end{table}

\section{LIMITATIONS AND FUTURE WORK}

It should be noted that to apply DyGEnc, keyframes must be extracted. In our robotic experiment, we perform this using uniform sampling at a one-second interval for a video. For long-term understanding, a more advanced system should be implemented to capture sparse key events; we do not attempt to solve this problem in the present work, leaving it for future research. However, DyGEnc already has two features to support long-term mode: the arbitrary shape of input graphs in the \textit{Q-Former} temporal encoder and a subgraph retrieval algorithm to reduce context size.

Also, despite achieving state-of-the-art results on the STAR and AGQA benchmarks, DyGEnc cannot yet be considered a foundational language model for graph encoding, as its capabilities are limited by the amount and diversity of training data with scene graph annotations compared to typical NLP tasks.

In future research, we plan to expand DyGEnc by exploring not only textual 2D scene graphs but also multimodal 3D scene graphs with temporal identification (object tracking). This requires a more complex dataset with graph and QA annotations, where each object is marked as an instance with a unique label. Unfortunately, the existing approach HyperGLM~\cite{nguyen2024hyperglm} has not published its codebase and data to the time of our research, thus necessitating a large amount of resource allocation for the creation of an open-source analog required first.

\section{CONCLUSION}

With DyGEnc, we advance the limits of dynamic scenes perception for robotics by integrating language models with a graph sequence encoder. The successful outcomes of our experiments on the complex STAR and AGQA, as well as on our real-life data, demonstrate the effectiveness of our approach, opening new avenues for a more comprehensive and flexible understanding and interaction with dynamic scenes. We hope that our code implementation will facilitate applications in real-world robotics projects that bridge the communication gap between humans and intelligent autonomous agents and robots.



\newpage

\bibliographystyle{IEEEtran}
\bibliography{IEEEabrv,bibliography}

\begin{thebibliography}{10}
\providecommand{\url}[1]{#1}
\csname url@rmstyle\endcsname
\providecommand{\newblock}{\relax}
\providecommand{\bibinfo}[2]{#2}
\providecommand\BIBentrySTDinterwordspacing{\spaceskip=0pt\relax}
\providecommand\BIBentryALTinterwordstretchfactor{4}
\providecommand\BIBentryALTinterwordspacing{\spaceskip=\fontdimen2\font plus
\BIBentryALTinterwordstretchfactor\fontdimen3\font minus \fontdimen4\font\relax}
\providecommand\BIBforeignlanguage[2]{{%
\expandafter\ifx\csname l@#1\endcsname\relax
\typeout{** WARNING: IEEEtran.bst: No hyphenation pattern has been}%
\typeout{** loaded for the language `#1'. Using the pattern for}%
\typeout{** the default language instead.}%
\else
\language=\csname l@#1\endcsname
\fi
#2}}

\bibitem{gu2024conceptgraphs}
Q.~Gu, A.~Kuwajerwala, S.~Morin, K.~M. Jatavallabhula, B.~Sen, A.~Agarwal, C.~Rivera, W.~Paul, K.~Ellis, R.~Chellappa, \emph{et~al.}, ``Conceptgraphs: Open-vocabulary 3d scene graphs for perception and planning,'' in \emph{2024 IEEE International Conference on Robotics and Automation (ICRA)}.\hskip 1em plus 0.5em minus 0.4em\relax IEEE, 2024, pp. 5021--5028.

\bibitem{linok2024beyond}
S.~Linok, T.~Zemskova, S.~Ladanova, R.~Titkov, and D.~Yudin, ``Beyond bare queries: Open-vocabulary object retrieval with 3d scene graph,'' \emph{arXiv e-prints}, pp. arXiv--2406, 2024.

\bibitem{takmaz2025search3d}
A.~Takmaz, A.~Delitzas, R.~W. Sumner, F.~Engelmann, J.~Wald, and F.~Tombari, ``Search3d: Hierarchical open-vocabulary 3d segmentation,'' \emph{IEEE Robotics and Automation Letters}, 2025.

\bibitem{werby2024hierarchical}
A.~Werby, C.~Huang, M.~B{\"u}chner, A.~Valada, and W.~Burgard, ``Hierarchical open-vocabulary 3d scene graphs for language-grounded robot navigation,'' in \emph{First Workshop on Vision-Language Models for Navigation and Manipulation at ICRA 2024}, 2024.

\bibitem{maggio2024clio}
D.~Maggio, Y.~Chang, N.~Hughes, M.~Trang, D.~Griffith, C.~Dougherty, E.~Cristofalo, L.~Schmid, and L.~Carlone, ``Clio: Real-time task-driven open-set 3d scene graphs,'' \emph{IEEE Robotics and Automation Letters}, 2024.

\bibitem{yang20234d}
J.~Yang, J.~Cen, W.~Peng, S.~Liu, F.~Hong, X.~Li, K.~Zhou, Q.~Chen, and Z.~Liu, ``4d panoptic scene graph generation,'' \emph{Advances in Neural Information Processing Systems}, vol.~36, pp. 69\,692--69\,705, 2023.

\bibitem{he2025g}
X.~He, Y.~Tian, Y.~Sun, N.~Chawla, T.~Laurent, Y.~LeCun, X.~Bresson, and B.~Hooi, ``G-retriever: Retrieval-augmented generation for textual graph understanding and question answering,'' \emph{Advances in Neural Information Processing Systems}, vol.~37, pp. 132\,876--132\,907, 2025.

\bibitem{perozzi2024let}
B.~Perozzi, B.~Fatemi, D.~Zelle, A.~Tsitsulin, M.~Kazemi, R.~Al-Rfou, and J.~Halcrow, ``Let your graph do the talking: Encoding structured data for llms,'' \emph{arXiv preprint arXiv:2402.05862}, 2024.

\bibitem{selvamcan}
K.~P. Selvam, P.~M. Phothilimthana, S.~Abu-El-Haija, B.~Perozzi, and M.~Brorsson, ``Can llms enhance performance prediction for deep learning models?''

\bibitem{wu2024star}
B.~Wu, S.~Yu, Z.~Chen, J.~B. Tenenbaum, and C.~Gan, ``Star: A benchmark for situated reasoning in real-world videos,'' \emph{arXiv preprint arXiv:2405.09711}, 2024.

\bibitem{grundemclaughlin2022agqa20updatedbenchmark}
\BIBentryALTinterwordspacing
M.~Grunde-McLaughlin, R.~Krishna, and M.~Agrawala, ``Agqa 2.0: An updated benchmark for compositional spatio-temporal reasoning,'' 2022. [Online]. Available: \url{https://arxiv.org/abs/2204.06105}
\BIBentrySTDinterwordspacing

\bibitem{krishna2017visual}
R.~Krishna, Y.~Zhu, O.~Groth, J.~Johnson, K.~Hata, J.~Kravitz, S.~Chen, Y.~Kalantidis, L.-J. Li, D.~A. Shamma, \emph{et~al.}, ``Visual genome: Connecting language and vision using crowdsourced dense image annotations,'' \emph{International journal of computer vision}, vol. 123, pp. 32--73, 2017.

\bibitem{hudson2019gqa}
D.~A. Hudson and C.~D. Manning, ``Gqa: A new dataset for real-world visual reasoning and compositional question answering,'' in \emph{Proceedings of the IEEE/CVF conference on computer vision and pattern recognition}, 2019, pp. 6700--6709.

\bibitem{yang2022panoptic}
J.~Yang, Y.~Z. Ang, Z.~Guo, K.~Zhou, W.~Zhang, and Z.~Liu, ``Panoptic scene graph generation,'' in \emph{European Conference on Computer Vision}.\hskip 1em plus 0.5em minus 0.4em\relax Springer, 2022, pp. 178--196.

\bibitem{ji2020action}
J.~Ji, R.~Krishna, L.~Fei-Fei, and J.~C. Niebles, ``Action genome: Actions as compositions of spatio-temporal scene graphs,'' in \emph{Proceedings of the IEEE/CVF conference on computer vision and pattern recognition}, 2020, pp. 10\,236--10\,247.

\bibitem{yang2023panoptic}
J.~Yang, W.~Peng, X.~Li, Z.~Guo, L.~Chen, B.~Li, Z.~Ma, K.~Zhou, W.~Zhang, C.~C. Loy, \emph{et~al.}, ``Panoptic video scene graph generation,'' in \emph{Proceedings of the IEEE/CVF Conference on Computer Vision and Pattern Recognition}, 2023, pp. 18\,675--18\,685.

\bibitem{im2024egtr}
J.~Im, J.~Nam, N.~Park, H.~Lee, and S.~Park, ``Egtr: Extracting graph from transformer for scene graph generation,'' in \emph{Proceedings of the IEEE/CVF Conference on Computer Vision and Pattern Recognition}, 2024, pp. 24\,229--24\,238.

\bibitem{cong2023reltr}
Y.~Cong, M.~Y. Yang, and B.~Rosenhahn, ``Reltr: Relation transformer for scene graph generation,'' \emph{IEEE Transactions on Pattern Analysis and Machine Intelligence}, vol.~45, no.~9, pp. 11\,169--11\,183, 2023.

\bibitem{wang2024oed}
G.~Wang, Z.~Li, Q.~Chen, and Y.~Liu, ``Oed: towards one-stage end-to-end dynamic scene graph generation,'' in \emph{Proceedings of the IEEE/CVF Conference on Computer Vision and Pattern Recognition}, 2024, pp. 27\,938--27\,947.

\bibitem{ravi2024sam}
N.~Ravi, V.~Gabeur, Y.-T. Hu, R.~Hu, C.~Ryali, T.~Ma, H.~Khedr, R.~R{\"a}dle, C.~Rolland, L.~Gustafson, \emph{et~al.}, ``Sam 2: Segment anything in images and videos,'' \emph{arXiv preprint arXiv:2408.00714}, 2024.

\bibitem{liu2024llavanext}
H.~Liu, C.~Li, Y.~Li, B.~Li, Y.~Zhang, S.~Shen, and Y.~J. Lee, ``Llavanext: Improved reasoning, ocr, and world knowledge,'' 2024.

\bibitem{cheng2024yolo}
T.~Cheng, L.~Song, Y.~Ge, W.~Liu, X.~Wang, and Y.~Shan, ``Yolo-world: Real-time open-vocabulary object detection,'' in \emph{Proceedings of the IEEE/CVF Conference on Computer Vision and Pattern Recognition}, 2024, pp. 16\,901--16\,911.

\bibitem{achiam2023gpt}
J.~Achiam, S.~Adler, S.~Agarwal, L.~Ahmad, I.~Akkaya, F.~L. Aleman, D.~Almeida, J.~Altenschmidt, S.~Altman, S.~Anadkat, \emph{et~al.}, ``Gpt-4 technical report,'' \emph{arXiv preprint arXiv:2303.08774}, 2023.

\bibitem{young2024yi}
A.~Young, B.~Chen, C.~Li, C.~Huang, G.~Zhang, G.~Zhang, G.~Wang, H.~Li, J.~Zhu, J.~Chen, \emph{et~al.}, ``Yi: Open foundation models by 01. ai,'' \emph{arXiv preprint arXiv:2403.04652}, 2024.

\bibitem{liu2024nvila}
Z.~Liu, L.~Zhu, B.~Shi, Z.~Zhang, Y.~Lou, S.~Yang, H.~Xi, S.~Cao, Y.~Gu, D.~Li, \emph{et~al.}, ``Nvila: Efficient frontier visual language models,'' \emph{arXiv preprint arXiv:2412.04468}, 2024.

\bibitem{bai2025qwen2}
S.~Bai, K.~Chen, X.~Liu, J.~Wang, W.~Ge, S.~Song, K.~Dang, P.~Wang, S.~Wang, J.~Tang, \emph{et~al.}, ``Qwen2. 5-vl technical report,'' \emph{arXiv preprint arXiv:2502.13923}, 2025.

\bibitem{GBC2024}
Y.-G. Hsieh, C.-Y. Hsieh, S.-Y. Yeh, L.~Béthune, H.~Pouransari, P.~K.~A. Vasu, C.-L. Li, R.~Krishna, O.~Tuzel, and M.~Cuturi, ``Graph-based captioning: Enhancing visual descriptions by interconnecting region captions,'' \emph{arXiv preprint arXiv:2407.06723}, 2024.

\bibitem{zhang2024provision}
J.~Zhang, L.~Xue, L.~Song, J.~Wang, W.~Huang, M.~Shu, A.~Yan, Z.~Ma, J.~C. Niebles, C.~Xiong, \emph{et~al.}, ``Provision: Programmatically scaling vision-centric instruction data for multimodal language models,'' \emph{arXiv preprint arXiv:2412.07012}, 2024.

\bibitem{kim2024llm4sgg}
K.~Kim, K.~Yoon, J.~Jeon, Y.~In, J.~Moon, D.~Kim, and C.~Park, ``Llm4sgg: large language models for weakly supervised scene graph generation,'' in \emph{Proceedings of the IEEE/CVF Conference on Computer Vision and Pattern Recognition}, 2024, pp. 28\,306--28\,316.

\bibitem{yang2024thinking}
J.~Yang, S.~Yang, A.~W. Gupta, R.~Han, L.~Fei-Fei, and S.~Xie, ``Thinking in space: How multimodal large language models see, remember, and recall spaces,'' \emph{arXiv preprint arXiv:2412.14171}, 2024.

\bibitem{zou2024seconds}
H.~Zou, T.~Luo, G.~Xie, F.~Lv, G.~Wang, J.~Chen, Z.~Wang, H.~Zhang, H.~Zhang, \emph{et~al.}, ``From seconds to hours: Reviewing multimodal large language models on comprehensive long video understanding,'' \emph{arXiv preprint arXiv:2409.18938}, 2024.

\bibitem{li2025visual}
Y.~Li, Z.~Lai, W.~Bao, Z.~Tan, A.~Dao, K.~Sui, J.~Shen, D.~Liu, H.~Liu, and Y.~Kong, ``Visual large language models for generalized and specialized applications,'' \emph{arXiv preprint arXiv:2501.02765}, 2025.

\bibitem{cherian20222}
A.~Cherian, C.~Hori, T.~K. Marks, and J.~Le~Roux, ``(2.5+ 1) d spatio-temporal scene graphs for video question answering,'' in \emph{Proceedings of the AAAI Conference on Artificial Intelligence}, vol.~36, no.~1, 2022, pp. 444--453.

\bibitem{rodin2024action}
I.~Rodin, A.~Furnari, K.~Min, S.~Tripathi, and G.~M. Farinella, ``Action scene graphs for long-form understanding of egocentric videos,'' in \emph{Proceedings of the IEEE/CVF Conference on Computer Vision and Pattern Recognition}, 2024, pp. 18\,622--18\,632.

\bibitem{nguyen2024hyperglm}
T.-T. Nguyen, P.~Nguyen, J.~Cothren, A.~Yilmaz, and K.~Luu, ``Hyperglm: Hypergraph for video scene graph generation and anticipation,'' \emph{arXiv preprint arXiv:2411.18042}, 2024.

\bibitem{qiu2024step}
H.~Qiu, M.~Gao, L.~Qian, K.~Pan, Q.~Yu, J.~Li, W.~Wang, S.~Tang, Y.~Zhuang, and T.-S. Chua, ``Step: Enhancing video-llms' compositional reasoning by spatio-temporal graph-guided self-training,'' \emph{arXiv preprint arXiv:2412.00161}, 2024.

\bibitem{warner2412smarter}
B.~Warner, A.~Chaffin, B.~Clavi{\'e}, O.~Weller, O.~Hallstrom, S.~Taghadouini, A.~Gallagher, R.~Biswas, F.~Ladhak, T.~Aarsen, \emph{et~al.}, ``Smarter, better, faster, longer: A modern bidirectional encoder for fast, memory efficient, and long context finetuning and inference,'' \emph{arXiv preprint arXiv.2412.13663}, 2024.

\bibitem{dwivedi2023benchmarking}
V.~P. Dwivedi, C.~K. Joshi, A.~T. Luu, T.~Laurent, Y.~Bengio, and X.~Bresson, ``Benchmarking graph neural networks,'' \emph{Journal of Machine Learning Research}, vol.~24, no.~43, pp. 1--48, 2023.

\bibitem{shi2020masked}
Y.~Shi, Z.~Huang, S.~Feng, H.~Zhong, W.~Wang, and Y.~Sun, ``Masked label prediction: Unified message passing model for semi-supervised classification,'' \emph{arXiv preprint arXiv:2009.03509}, 2020.

\bibitem{su2024roformer}
J.~Su, M.~Ahmed, Y.~Lu, S.~Pan, W.~Bo, and Y.~Liu, ``Roformer: Enhanced transformer with rotary position embedding,'' \emph{Neurocomputing}, vol. 568, p. 127063, 2024.

\bibitem{li2023blip}
J.~Li, D.~Li, S.~Savarese, and S.~Hoi, ``Blip-2: Bootstrapping language-image pre-training with frozen image encoders and large language models,'' in \emph{International conference on machine learning}.\hskip 1em plus 0.5em minus 0.4em\relax PMLR, 2023, pp. 19\,730--19\,742.

\bibitem{grunde2021agqa}
M.~Grunde-McLaughlin, R.~Krishna, and M.~Agrawala, ``Agqa: A benchmark for compositional spatio-temporal reasoning,'' in \emph{Proceedings of the IEEE/CVF Conference on Computer Vision and Pattern Recognition}, 2021, pp. 11\,287--11\,297.

\bibitem{hu2022lora}
E.~J. Hu, Y.~Shen, P.~Wallis, Z.~Allen-Zhu, Y.~Li, S.~Wang, L.~Wang, W.~Chen, \emph{et~al.}, ``Lora: Low-rank adaptation of large language models.'' \emph{ICLR}, vol.~1, no.~2, p.~3, 2022.

\bibitem{dubey2024llama}
A.~Dubey, A.~Jauhri, A.~Pandey, A.~Kadian, A.~Al-Dahle, A.~Letman, A.~Mathur, A.~Schelten, A.~Yang, A.~Fan, \emph{et~al.}, ``The llama 3 herd of models,'' \emph{arXiv preprint arXiv:2407.21783}, 2024.

\bibitem{cong2023we}
W.~Cong, S.~Zhang, J.~Kang, B.~Yuan, H.~Wu, X.~Zhou, H.~Tong, and M.~Mahdavi, ``Do we really need complicated model architectures for temporal networks?'' \emph{arXiv preprint arXiv:2302.11636}, 2023.

\bibitem{vaswani2017attention}
A.~Vaswani, N.~Shazeer, N.~Parmar, J.~Uszkoreit, L.~Jones, A.~N. Gomez, {\L}.~Kaiser, and I.~Polosukhin, ``Attention is all you need,'' \emph{Advances in neural information processing systems}, vol.~30, 2017.

\bibitem{romero2024question}
D.~Romero and T.~Solorio, ``Question-instructed visual descriptions for zero-shot video question answering,'' \emph{arXiv preprint arXiv:2402.10698}, 2024.

\bibitem{gao2023mist}
D.~Gao, L.~Zhou, L.~Ji, L.~Zhu, Y.~Yang, and M.~Z. Shou, ``Mist: Multi-modal iterative spatial-temporal transformer for long-form video question answering,'' in \emph{Proceedings of the IEEE/CVF conference on computer vision and pattern recognition}, 2023, pp. 14\,773--14\,783.

\bibitem{yu2023self}
S.~Yu, J.~Cho, P.~Yadav, and M.~Bansal, ``Self-chained image-language model for video localization and question answering,'' \emph{Advances in Neural Information Processing Systems}, vol.~36, pp. 76\,749--76\,771, 2023.

\bibitem{wang2024vila}
X.~Wang, J.~Liang, C.-K. Wang, K.~Deng, Y.~Lou, M.~C. Lin, and S.~Yang, ``Vila: Efficient video-language alignment for video question answering,'' in \emph{European Conference on Computer Vision}.\hskip 1em plus 0.5em minus 0.4em\relax Springer, 2024, pp. 186--204.

\bibitem{liang2024end}
J.~Liang, X.~Meng, Y.~Wang, C.~Liu, Q.~Liu, and D.~Zhao, ``End-to-end video question answering with frame scoring mechanisms and adaptive sampling,'' \emph{arXiv preprint arXiv:2407.15047}, 2024.

\bibitem{bhattacharyya2023look}
A.~Bhattacharyya, S.~Panchal, M.~Lee, R.~Pourreza, P.~Madan, and R.~Memisevic, ``Look, remember and reason: Grounded reasoning in videos with language models,'' \emph{arXiv preprint arXiv:2306.17778}, 2023.

\bibitem{bai2023glance}
Z.~Bai, R.~Wang, and X.~Chen, ``Glance and focus: Memory prompting for multi-event video question answering,'' \emph{Advances in Neural Information Processing Systems}, vol.~36, pp. 34\,247--34\,259, 2023.

\bibitem{jaiswal2025learning}
S.~Jaiswal, D.~Roy, B.~Fernando, and C.~Tan, ``Learning to reason iteratively and parallelly for complex visual reasoning scenarios,'' \emph{Advances in Neural Information Processing Systems}, vol.~37, pp. 137\,965--137\,998, 2025.

\bibitem{wang2024efficient}
Y.~Wang, Y.~Wang, P.~Wu, J.~Liang, D.~Zhao, Y.~Liu, and Z.~Zheng, ``Efficient temporal extrapolation of multimodal large language models with temporal grounding bridge,'' \emph{arXiv preprint arXiv:2402.16050}, 2024.

\bibitem{lee2022learning}
H.-Y. Lee, H.-T. Su, B.-C. Tsai, T.-H. Wu, J.-F. Yeh, and W.~H. Hsu, ``Learning fine-grained visual understanding for video question answering via decoupling spatial-temporal modeling,'' \emph{arXiv preprint arXiv:2210.03941}, 2022.

\bibitem{he2024g}
X.~He, Y.~Tian, Y.~Sun, N.~V. Chawla, T.~Laurent, Y.~LeCun, X.~Bresson, and B.~Hooi, ``G-retriever: Retrieval-augmented generation for textual graph understanding and question answering,'' \emph{arXiv preprint arXiv:2402.07630}, 2024.

\bibitem{bienstock1993note_pcst}
D.~Bienstock, M.~X. Goemans, D.~Simchi-Levi, and D.~Williamson, ``A note on the prize collecting traveling salesman problem,'' \emph{Mathematical programming}, vol.~59, no. 1-3, pp. 413--420, 1993.

\bibitem{li2023factual}
Z.~Li, Y.~Chai, T.~Y. Zhuo, L.~Qu, G.~Haffari, F.~Li, D.~Ji, and Q.~H. Tran, ``Factual: A benchmark for faithful and consistent textual scene graph parsing,'' \emph{arXiv preprint arXiv:2305.17497}, 2023.

\end{thebibliography}

\end{document}